%% file: IEEE-conference-template-062824.tex
\documentclass[9pt,conference]{IEEEtran}
\IEEEoverridecommandlockouts

\usepackage{cite}
\usepackage{amsmath,amssymb,amsfonts}
\usepackage{algorithmic}
\usepackage{graphicx}
\usepackage{textcomp}
\usepackage[dvipsnames]{xcolor}
\usepackage{adjustbox}
\usepackage{multirow}
\usepackage{booktabs}
\usepackage{pifont}
\usepackage{bbding}
\usepackage{colortbl}
\usepackage{orcidlink}
\usepackage{hyperref}
\usepackage[capitalize]{cleveref}
\crefname{section}{Sec.}{Secs.}
\Crefname{section}{Section}{Sections}
\Crefname{table}{Table}{Tables}
\crefname{table}{Table}{Tables}
\Crefname{figure}{Figure}{Figures}
\crefname{figure}{Fig.}{Figs.}
\Crefname{equation}{Equation}{Equations}
\crefname{equation}{Eq.}{Eqs.}
\crefname{algocf}{alg.}{algs.}
\Crefname{algocf}{Algorithm}{Algorithms}

\def\BibTeX{{\rm B\kern-.05em{\sc i\kern-.025em b}\kern-.08em
    T\kern-.1667em\lower.7ex\hbox{E}\kern-.125emX}}
\begin{document}
\title{Underwater Image Restoration via Polymorphic \\ Large Kernel CNNs
\thanks{* These authors contribute equally to this work.}
\thanks{$^\dag$ Corresponding authors.}
}

\author{
\IEEEauthorblockN{
Xiaojiao Guo\textsuperscript{16*} \qquad 
Yihang Dong\textsuperscript{2*} \qquad 
Xuhang Chen\textsuperscript{123\orcidlink{0000-0001-6000-3914}} \qquad 
Weiwen Chen\textsuperscript{15} \qquad 
Zimeng Li\textsuperscript{4$\dag$} \\
FuChen Zheng\textsuperscript{12} \qquad
Chi-Man Pun\textsuperscript{1$\dag$}
}
\IEEEauthorblockA{
\textsuperscript{1}University of Macau \qquad
\textsuperscript{2}Shenzhen Institute of Advanced Technology, Chinese Academy of Sciences \qquad
\textsuperscript{3}Huizhou Univeristy \\
\textsuperscript{4}Shenzhen Polytechnic University \qquad
\textsuperscript{5}The Hong Kong University of Science and Technology (Guangzhou) \qquad
\textsuperscript{6}Baoshan Univeristy
}
}

\maketitle

\begin{abstract}
Underwater Image Restoration (UIR) remains a challenging task in computer vision due to the complex degradation of images in underwater environments. While recent approaches have leveraged various deep learning techniques, including Transformers and complex, parameter-heavy models to achieve significant improvements in restoration effects, we demonstrate that pure CNN architectures with lightweight parameters can achieve comparable results.
In this paper, we introduce UIR-PolyKernel, a novel method for underwater image restoration that leverages Polymorphic Large Kernel CNNs. Our approach uniquely combines large kernel convolutions of diverse sizes and shapes to effectively capture long-range dependencies within underwater imagery.
Additionally, we introduce a Hybrid Domain Attention module that integrates frequency and spatial domain attention mechanisms to enhance feature importance. By leveraging the frequency domain, we can capture hidden features that may not be perceptible to humans but are crucial for identifying patterns in both underwater and on-air images. This approach enhances the generalization and robustness of our UIR model.
Extensive experiments on benchmark datasets demonstrate that UIR-PolyKernel achieves state-of-the-art performance in underwater image restoration tasks, both quantitatively and qualitatively. Our results show that well-designed pure CNN architectures can effectively compete with more complex models, offering a balance between performance and computational efficiency. This work provides new insights into the potential of CNN-based approaches for challenging image restoration tasks in underwater environments. The code is available at \href{https://github.com/CXH-Research/UIR-PolyKernel}{https://github.com/CXH-Research/UIR-PolyKernel}.
\end{abstract}

\begin{IEEEkeywords}
Underwater image restoration, image enhancement, large kernel CNNs, hybrid domain, frequency domain.
\end{IEEEkeywords}

\section{Introduction}
Underwater image restoration (UIR) is critical for a wide range of applications, but images captured underwater often suffer from severe degradation due to the complex optical properties of water. This degradation, characterized by reduced visibility, color distortion, and blurred details, hinders the effectiveness of underwater imaging systems. Traditional UIR methods, such as UNTV~\cite{xie2021variational}, MLLE~\cite{zhang2022underwater}, ROP~\cite{liu2022rank}, ADPCC~\cite{zhou2023underwater}, and WWPF~\cite{zhang2023underwater}, have laid a foundation for addressing these challenges but often struggle to generalize across diverse water types and complex underwater scenes due to their reliance on pre-defined models or specific assumptions.

Deep learning has emerged as a powerful approach for UIR, offering greater flexibility and the ability to learn complex degradation models directly from data~\cite{zhang1,zhang6}. This has led to the development of various successful CNN-based methods like Ucolor~\cite{li2021underwater}, CLUIE-Net~\cite{li2022beyond}, and SFGNet~\cite{zhao2024toward}, as well as GAN-based frameworks like TACL~\cite{liu2022twin}, UIE-WD~\cite{ma2022wavelet}, PUGAN~\cite{cong2023pugan}, and TUDA~\cite{wang2023domain}. More recently, Transformer-based architectures, such as U-Transformer~\cite{peng2023u} and URSCT~\cite{ren2022reinforced}, have shown promise due to their ability to capture long-range dependencies within images. Addressing the scarcity of labeled data in underwater environments, researchers have explored semi-supervised learning (e.g., Semi-UIR~\cite{huang2023contrastive}) and unsupervised learning (e.g., USUIR~\cite{fu2022unsupervised}). Additionally, hybrid methods like GUPDM~\cite{mu2023generalized} integrate deep learning with prior knowledge, demonstrating the potential of physics-guided approaches. However, those high performance methods among them often come with sophisticated architecture and high computational costs, limiting their practical application in resource-constrained scenarios~\cite{zhu2024test,xu2024muraldiff, zhang8,li2024cross,he2024residual,li2022nndf}.

The success of Transformer-based models can be attributed to their ability to capture long-range dependencies within images~\cite{zhang2,liu2024forgeryttt,li2024high-fidelity,zhang2024seeing,liu2024dh,li2023high-resolution,guo2024dual-hybrid,liu2023explicit,zhang10,zhang11}. However, their self-attention mechanism has a high computational complexity, making them expensive for high-resolution images or real-time applications~\cite{yuan2024instance,zhang9,liu2023coordfill,liu2024depth,he2024generalized}. Recent research has shown that large kernel convolutions~\cite{guo2023visual, ding2022scaling, cui2024omni} can effectively mimic the long-range dependency capture of Transformers with lower computational overhead.

To leverage the local detail depiction capabilities of CNNs and the long-range relationship capturing power similar to that of Transformers while maintaining efficiency, we make the following key contributions:
\begin{itemize}
\item We propose UIR-PolyKernel, a lightweight and computationally efficient pure CNN architecture that achieves state-of-the-art performance in underwater image restoration by integrating polymorphic large kernel convolutions of diverse sizes, shapes, and depths.
\item We introduce a Hybrid Domain Attention module that integrates frequency and spatial domain attention mechanisms to enhance feature importance, capturing hidden features that may not be perceptible in the spatial domain and enhancing the network's ability to restore fine-grained details.
\item Through extensive experiments on benchmark datasets, we demonstrate that UIR-PolyKernel consistently outperforms existing methods, including more complex models, challenging the notion that complex, heavy models are necessary for this task.
\end{itemize}

\input{figtex/overall_arch}

\section{Methodology}

We present UIR-PolyKernel, an encoder-bottleneck-decoder network for underwater image restoration as shown in \cref{fig:overall}. The network leverages large kernel convolutions applied to downsampled features to efficiently capture long-range dependencies within the image. Large Kernel Attention modules (with medium kernel size) are strategically positioned near the bottleneck to progressively expand the receptive field. The bottleneck itself incorporates a Composite Shape Convolution module, which integrates heterotypic convolutions employing extremely large kernels ($31\times1$ Horizontal Strip Convolution, $1\times31$ Vertical Strip Convolution, and $31\times31$ Square Kernel Convolution). This combination allows for comprehensive capture of global dependencies, effectively modeling long-range relationships across the entire feature map. Finally, a hybrid domain attention mechanism operating on the highest resolution input/output feature maps preserves fine-grained details crucial for high-fidelity image restoration. The following subsections provide a detailed description of each module, starting with the bottleneck and moving towards the finest encoder/decoder scales.

\subsection{Composite Shape Convolution (CSC) Module}
The feature maps inputted to the bottleneck come from the encoder's final module and are at a reduced resolution of one-fourth the original input; with our network primarily trained on $256 \times 256$ image patches, this results in $64 \times 64$ feature maps. Inspired by OKNet~\cite{cui2024omni}, our initial exploration involved employing extremely large kernel convolutions (e.g., $63 \times 63$) at this stage to capture global contextual information, crucial for restoring spatially coherent features often degraded in underwater images. However, such large kernels incur significant computational costs. To strike a balance between receptive field size and computational efficiency, we experimented with different kernel sizes ($15$, $31$, and $63$), ultimately selecting $31\times31$ kernels.

To further mitigate the computational burden of square-shaped large kernels and capture multi-scale features relevant to varying degrees of degradation in underwater scenes, we take inspiration from OKNet and LSKA~\cite{lau2024large} and adopt strip-shaped convolutions. The core part of the CSC module is the parallel composition of four depth-wise polymorphic kernel convolutions: $31 \times 1$ (horizontal), $1 \times 31$ (vertical), $31 \times 31$ (square) and $1 \times 1$ (point-wise). It can be formulated as:
\begin{equation}
\begin{split}
\operatorname{CSC_{core}} (\mathbf{F} )=&{Conv}^{dw} _{31 \times 1}(\mathbf{F} ) + {Conv}^{dw} _{1 \times 31}(\mathbf{F} )\\ &+ {Conv}^{dw} _{31 \times 31}(\mathbf{F} ) + {Conv}^{dw} _{1 \times 1}(\mathbf{F} ),
\end{split}
\label{eq:CSC}
\end{equation}
where $\mathbf{F}$ represents the input feature map, and ${Conv}^{dw} _{k \times k}$ denotes a depth-wise convolution with a kernel size of $k \times k$. The strip-shaped convolutions in the CSC module allow the network to capture anisotropic features common in underwater scenes, such as the varying degrees of attenuation and scattering along different spatial dimensions. The square-shaped convolution ensures that the module can still capture isotropic features when necessary. The $1 \times 1$ point-wise depth-wise convolution calibrates pixel-wise weights directly, enhancing the module's ability to adapt to local variations in the image.

Since depth-wise convolutions only process relationships within the same channel, an additional $1 \times 1$ point-wise convolution follows this parallel architecture to calibrate inter-channel dependencies. This combination of diverse, extremely large kernel shapes and efficient depth-wise convolutions forms our proposed CSC Module. This module effectively captures long-range relationships without excessively increasing computational demands, which is particularly important at the bottleneck stage where global context is crucial for restoring visually coherent underwater images.

\subsection{Large Kernel Attention (LKA) Module}
While the CSC Module excels at capturing global context, enhancing multi-scale feature representation requires attention to medium-range relationships. To address this, we introduce Large Kernel Attention (LKA) Modules at two levels within the encoder-decoder structure, positioned symmetrically in both the encoder and decoder. These modules operate on feature maps downsampled to half and one-quarter of the original resolution, respectively.

Inspired by VAN~\cite{guo2023visual}, each LKA Module employs a sequential arrangement of a $5 \times 5$ depth-wise convolution, a $7 \times 7$ depth-wise dilated convolution (dilation rate 3), and an $1 \times 1$ convolution. This configuration effectively simulates a $19 \times 19$ convolution, enabling the network to capture medium-range relationships crucial for reconstructing underwater images often exhibiting degradation and feature correlations at these scales. The resulting attention maps are then multiplied element-wise with the input feature maps, guiding the network to focus on relevant medium-scale features. This mechanism bridges the gap between fine-grained details and the global context, fostering a hierarchical understanding of the scene essential for high-quality underwater image restoration.

\subsection{Hybrid Domain Attention (HDA) Module}
While the CSC and LKA modules effectively address large-scale degradation and color distortion by capturing global and medium-range features, preserving fine textures and details requires attention at the finest level of detail. To this end, we introduce the Hybrid Domain Attention (HDA) Module, strategically positioned at the beginning of the encoder and the end of the decoder, respectively, operating on feature maps that retain the original input resolution.

\input{tabletex/comp}

Recognizing that frequency domain analysis offers a complementary perspective on image features, our HDA Module leverages both spatial and frequency domain information. As illustrated in \cref{fig:overall}, the HDA Module contains two sequential blocks: the Frequency-Domain Pixel Attention (FDPA) block and the Spatial-Domain Channel Attention (SDCA) block. In the FDPA block, the input feature maps first undergo two $1 \times 1$ convolution branches. One branch is transformed into the frequency domain using a Fast Fourier Transform (FFT), while the other branch remains in the spatial domain. The spatial domain features from the second branch are then multiplied element-wise with the frequency domain representation from the first branch, effectively integrating attention from both domains. After applying an inverse FFT, this frequency domain attention map is combined with the original input features through a weighted addition. The entire process of FDPA can be formulated as:
\begin{equation}
\operatorname{FDPA} (\mathbf{F} ) = \alpha \cdot \operatorname{IFFT} (\operatorname{FFT} (Conv1(\mathbf{F} )) \odot Conv1 (\mathbf{F})) + \beta \cdot \mathbf{F},
\label{eq:FDPA}
\end{equation}
where $\mathbf{F}$ represents the input features, $Conv1$ denotes an $1 \times 1$ convolution, $\odot$ represents element-wise multiplication, and $\alpha$ and $\beta$ are learnable parameters that control the contribution of the frequency domain attention and the original input features, respectively.

To further refine the feature representation, we incorporate channel-wise attention in the SDCA block. Global average pooling is applied to the output of the previous step, generating a channel-wise descriptor. This descriptor is then used to spatially calibrate the feature map, emphasizing channels that contribute most significantly to the restoration task. This combination of spatial and frequency domain attention, along with channel-wise refinement, allows the HDA Module to effectively capture and enhance fine details, complementing the coarser-level processing of the CSC and LKA modules.

\subsection{Loss Function}

The primary objective during model training is to minimize the difference between the restored output image and the reference target image. It is important to note that the reference target, while consisting of high-quality results from state-of-the-art methods, does not represent the true ground truth. Our goal is to surpass these existing methods and establish a new benchmark in underwater image restoration. To achieve this, we employ a composite loss function:
\begin{equation}
\mathcal{L} = w_1 \cdot \mathcal{L}_{p} + w_2 \cdot \mathcal{L}_{s} + w_3 \cdot \mathcal{L}_{u},
\label{eq:Loss}
\end{equation}
where $\mathcal{L}_{p}$ denotes the pixel-level similarity loss, specifically the SmoothL1Loss, $\mathcal{L}_{s}$ represents the structural similarity loss, which is equivalent to $1 - \text{SSIM}$, and $\mathcal{L}_{u}$ pertains to the underwater image quality loss, equivalent to $1-\text{UCIQE}$. The weights $w_1$, $w_2$, and $w_3$, which have been optimized through experimental procedures, are assigned values of $1$, $0.2$, and $0.01$, respectively.

\input{figtex/compare}

\section{Experiment}
\subsection{Datasets and Evaluation Metrics}

Underwater image datasets are generally divided into synthetic and real-world types, with models trained on synthetic data often struggling to adapt to real scenes. To address this, we selected training data from three high-quality real-world datasets: UIEB~\cite{li2019underwater}, EUVP~\cite{islam2020fast}, and LSUI~\cite{peng2023u}, which offer diverse and high-quality underwater images. Specifically, we randomly sampled 800 images from UIEB, 2000 from EUVP, and 2000 from LSUI for training.

For a comprehensive evaluation of the performance and generalization of our method, we compiled paired test sets consisting of 90 UIEB images, 200 EUVP samples, and 200 LSUI samples, ensuring no overlap with the training set. Additionally, to test the generalizability to non-trained domains, we included 200 unpaired test images from RUIE~\cite{liu2020real}. This diverse set of test data allows for a comprehensive evaluation of the methods' performance and generalization capabilities across a wide range of underwater images with varying levels of difficulty.

We employ four widely recognized metrics, including reference metrics PSNR, SSIM, and LPIPS, and the non-reference metric UCIQE. PSNR and SSIM assess the fidelity and structural similarity of the pixels, while LPIPS evaluates the perceptual similarity based on human perception. UCIQE, specifically designed for underwater images, considers reasonable imaging parameters such as chroma, saturation, and contrast. These diverse metrics provide a thorough and balanced assessment of each method's performance in the underwater image restoration task.

\subsection{Implementation Details}
Experiments use PyTorch on Ubuntu with an NVIDIA RTX 4090 GPU. A batch size of 16 is employed, with initial and minimal learning rates of $2\times10^{-4}$ and $1\times10^{-6}$, respectively. The AdamW optimizer and Cosine Annealing Learning Rate Scheduler are applied. Training lasts 500 epochs with images resized to $256 \times 256$ pixels, using data augmentation like cropping, flipping, rotation, transposition, and scaling.

\subsection{Comparisons with State-of-the-arts}
\cref{tab:comp} presents the quantitative results of our experiments. For a fair comparison, all deep learning methods were retrained on the same training datasets as our method, with settings consistent with their original papers. 
UIR-PolyKernel achieved the highest scores in most referenced metrics, with the exception of ranking second behind Semi-UIR~\cite{huang2023contrastive} in PSNR on the UIEB dataset. In all other results, our method leads other approaches by a large margin. Although UCIQE tends to favor traditional methods, which often exhibit lower visual quality compared to deep learning approaches, we limit our UCIQE comparison to the deep learning methods. Although not absolutely reliable, the UCIQE metric still provides a certain level of reference value. As observed in \cref{tab:comp}, our method achieves the highest UCIQE score on all test sets compared to other deep learning methods.
This superior performance demonstrates the effectiveness of our approach in restoring underwater images and its ability to generalize well across different underwater environments.

To provide a comprehensive evaluation of the methods, we also report the computational cost in terms of MACs (Multiply-Accumulate Operations) and the number of parameters (Params) for each model in \cref{tab:comp}. Our method achieves the best balance between computational cost and evaluation metrics.

\Cref{fig:comp} provides a visual comparison of our method against two leading traditional approaches (ADPCC~\cite{zhou2023underwater}, WWPF~\cite{zhang2023underwater}) and the three top-performing deep learning methods (GUPDM~\cite{mu2023generalized}, Semi-UIR~\cite{huang2023contrastive}, URSCT~\cite{ren2022reinforced}). As illustrated in the figure, UIR-PolyKernel effectively addresses various underwater image degradation issues, including color distortion, blurriness, and low contrast. The restored images exhibit vibrant colors, sharp details, and improved overall visual quality, closely resembling the target images. This visual comparison further confirms the effectiveness of our proposed method in producing high-quality underwater image restoration results.

\subsection{Ablation Studies}
\input{tabletex/ablation}
We conduct ablation studies to validate each component of UIR-PolyKernel. \cref{tab:ab} shows results averaged across the UIEB, EUVP, and LSUI datasets. We constructed the model starting from a backbone UNet with the same overall architecture and number of modules as our full model, but with a higher parameter count of 1.985M compared to our full model's 1.837M. This ensures that the effectiveness of our method is not solely due to an increase in parameter count.
 
We progressively incorporated our proposed modules into the backbone U-Net, replacing the corresponding original modules at each step. First, we conducted two separate experiments: one adding the CSC module at the bottleneck and another incorporating the LKA module at two downsampled layers. Both modifications led to improvements across all metrics, highlighting the individual benefits of global context (CSC) and capturing medium-range relationships (LKA). Integrating both CSC and LKA modules into the backbone yielded even greater improvements, particularly in PSNR, demonstrating their complementary advantages in capturing multi-scale features. Next, we integrated the SDCA module into the HDA module, resulting in additional performance gains and highlighting the effectiveness of spatial channel-wise attention. Finally, adding the FDPA module to the HDA module led to the most significant performance gains, emphasizing the importance of leveraging attention in the spatial and frequency domains at the finest detail level.

These ablation studies demonstrate that each component in UIR-PolyKernel contributes to its superior performance. The combination of polymorphic large kernel convolutions and hybrid domain attention effectively addresses the challenges inherent in underwater image restoration.

\section{Conclusion}
This paper introduces UIR-PolyKernel, a novel underwater image restoration method using polymorphic large kernel CNNs. Our approach combines diverse large kernel convolutions within an encoder-bottleneck-decoder architecture, capturing global and medium-range dependencies. A hybrid domain attention module enhances fine texture preservation. Experiments demonstrate state-of-the-art performance on benchmark datasets, surpassing existing methods in both quantitative metrics and qualitative assessments. UIR-PolyKernel showcases the effectiveness of well-designed CNN architectures, offering a balance between performance and computational efficiency suitable for real-world applications.

\section*{Acknowledgment}
This work was supported in part by the University of Macau under Grants MYRG-GRG2023-00131-FST and MYRG-GRG2024-00065-FST-UMDF, in part by the Science and Technology Development Fund, Macau SAR, under Grants 0141/2023/RIA2 and 0193/2023/RIA3, in part by the National Natural Science Foundation of China under Grant 62272313.

\bibliographystyle{IEEEtran}
\bibliography{ref}

\end{document}

%% file: figtex/overall_arch.tex
\begin{figure*}[ht]
\centering
\includegraphics[width=\textwidth]{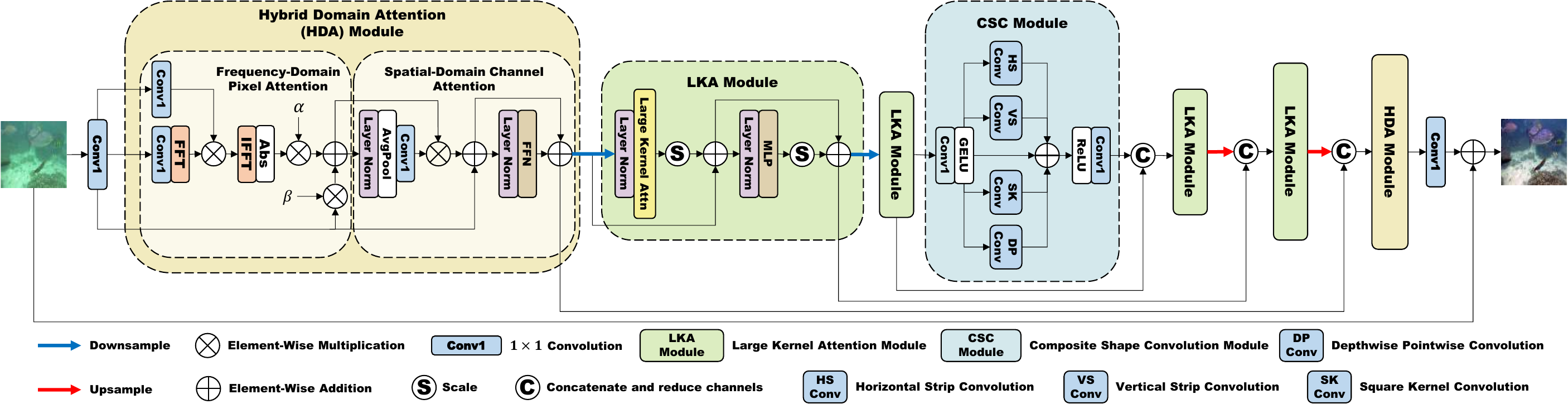}
\caption{The overall architecture of the proposed UIR-PolyKernel model. This model utilizes three resolution levels of feature maps, generated by two pairs of downsampling and upsampling operations. Each downsampling operation halves the spatial dimensions of the feature map while doubling the number of channels, with upsampling operations performing the inverse. The initial number of channels in the first level of feature maps is 36.
}
\label{fig:overall}
\end{figure*}

%% file: tabletex/comp.tex
\begin{table*}[ht]
\caption{Quantitative comparison on the paired test sets UIEB, EUVP, and LSUI, and the unpaired test set RUIE.}
\label{tab:comp}
\adjustbox{width=\linewidth}{
\begin{tabular}{l|rr|cccc|cccc|cccc|c}
\toprule
 & \multicolumn{2}{c|}{Computational Cost} & \multicolumn{4}{c|}{UIEB} & \multicolumn{4}{c|}{EUVP} & \multicolumn{4}{c|}{LSUI} & RUIE  \\ 
 \cmidrule(l){2-16} 
                        \multirow{-2}{*}{Method} & MACs(G)           & Params(M)           & PSNR     & SSIM    & LPIPS   & UCIQE  & PSNR     & SSIM    & LPIPS   & UCIQE   & PSNR     & SSIM    & LPIPS   & UCIQE   & UCIQE \\ \midrule
UNTV~\cite{xie2021variational}                     & \multicolumn{1}{c}{-}               & \multicolumn{1}{c|}{-}                 & 16.46   & 0.669   & 0.420   & 0.436  & 17.63   & 0.611   & 0.335   & \underline{0.481}   & 18.36   & 0.660   & 0.376   & 0.449   & 0.430 \\
MLLE~\cite{zhang2022underwater}                    & \multicolumn{1}{c}{-}               & \multicolumn{1}{c|}{-}                 & 18.74   & 0.814   & 0.234   & 0.444  & 15.14   & 0.633   & 0.323   & 0.474   & 17.87   & 0.730   & 0.278   & \underline{0.457}   & 0.434 \\
ROP~\cite{liu2022rank}                     & \multicolumn{1}{c}{-}               & \multicolumn{1}{c|}{-}                 & 18.48   & 0.849   & 0.209   & 0.454  & 15.34   & 0.714   & 0.343   & 0.441   & 17.38   & 0.806   & 0.281   & 0.448   & \textbf{0.461} \\
ADPCC~\cite{zhou2023underwater}                   & \multicolumn{1}{c}{-}               & \multicolumn{1}{c|}{-}                 & 17.33   & 0.819   & 0.219   & \textbf{0.485}  & 15.20   & 0.692   & 0.349   & \textbf{0.495}   & 16.20   & 0.763   & 0.299   & \textbf{0.481}   & \underline{0.447} \\
WWPF~\cite{zhang2023underwater}                    & \multicolumn{1}{c}{-}               & \multicolumn{1}{c|}{-}                 & 18.60   & 0.822   & 0.218   & 0.446  & 15.95   & 0.648   & 0.337   & 0.467   & 17.90   & 0.739   & 0.283   & 0.456   & 0.442 \\ \midrule
Ucolor~\cite{li2021underwater}                  & 1002.00           & 105.51              & 18.37   & 0.814   & 0.221   & 0.370  & 23.72   & 0.828   & 0.205   & 0.408   & 21.30   & 0.821   & 0.225   & 0.362   & 0.256 \\
CLUIE-Net~\cite{li2022beyond}               & 31.13             & 13.40               & 19.95   & 0.874   & 0.168   & 0.403  & 24.85   & 0.844   & 0.186   & 0.415   & 23.57   & 0.864   & 0.175   & 0.396   & 0.331 \\
TACL~\cite{liu2022twin}                    & 120.03            & 28.29               & 19.83   & 0.761   & 0.222   & 0.427  & 20.99   & 0.782   & 0.213   & 0.440   & 22.97   & 0.828   & 0.176   & 0.433   & 0.419 \\
UIE-WD~\cite{ma2022wavelet}                 & 51.38             & 14.49               & 20.28   & 0.848   & 0.198   & 0.407  & 17.80   & 0.760   & 0.292   & 0.440   & 19.23   & 0.803   & 0.284   & 0.410   & 0.343 \\
URSCT~\cite{ren2022reinforced}                   & 18.11             & 11.26               & 22.77   & 0.915   & 0.120   & 0.432  & \underline{25.74}   & \underline{0.855}   & 0.180   & 0.425   & \underline{25.87}   & \underline{0.883}   & \underline{0.146}   & 0.417   & 0.384 \\
USUIR~\cite{fu2022unsupervised}                   & 14.81             & 0.23                & 22.48   & 0.907   & 0.124   & 0.427  & 21.94   & 0.810   & 0.239   & 0.408   & 23.75   & 0.860   & 0.184   & 0.412   & 0.390 \\
GUPDM~\cite{mu2023generalized}                   & 95.80             & 1.49                & 22.13   & 0.903   & 0.131   & 0.427  & 24.79   & 0.847   & 0.184   & 0.431   & 25.33   & 0.877   & 0.150   & 0.420   & 0.380 \\
PUGAN~\cite{cong2023pugan}                   & 75.40             & 101.19              & 20.52   & 0.812   & 0.216   & 0.418  & 22.58   & 0.820   & 0.212   & 0.425   & 23.14   & 0.836   & 0.216   & 0.413   & 0.369 \\
Semi-UIR~\cite{huang2023contrastive}                & 72.88             & 3.31                & \textbf{23.64}   & 0.888   & 0.120   & 0.428  & 24.59   & 0.821   & \underline{0.172}   & 0.424   & 25.40   & 0.843   & 0.160   & 0.419   & 0.385 \\
TUDA~\cite{wang2023domain}                    & 85.43             & 2.73                & 22.72   & 0.915   & 0.118   & 0.429  & 23.73   & 0.843   & 0.207   & 0.415   & 25.52   & 0.878   & 0.154   & 0.417   & 0.392 \\
U-Transformer~\cite{peng2023u}           & 2.98              & 22.82               & 20.75   & 0.810   & 0.228   & 0.434  & 24.99   & 0.829   & 0.238   & 0.434   & 25.15   & 0.838   & 0.221   & 0.427   & 0.409 \\
SFGNet~\cite{zhao2024toward}                  & 81.58             & 1.30                & 19.57   & 0.685   & 0.214   & 0.432  & 22.68   & 0.585   & 0.221   & 0.442   & 22.71   & 0.653   & 0.204   & 0.426   & 0.390 \\
Ours                    & 13.67             & 1.84                & \underline{23.52}   & \textbf{0.925}   & \textbf{0.105}   & \underline{0.458}  & \textbf{26.42}   & \textbf{0.866}   & \textbf{0.154}   & 0.449   & \textbf{26.55}   & \textbf{0.888}   & \textbf{0.125}   & 0.451   & 0.440 \\ \bottomrule
\end{tabular}
}
\end{table*}

%% file: figtex/compare.tex
\begin{figure*}[ht]
    \centering
    \begin{minipage}[b]{1\linewidth}
        \begin{minipage}[b]{.12\linewidth}
            \centering
            \includegraphics[width=\linewidth]{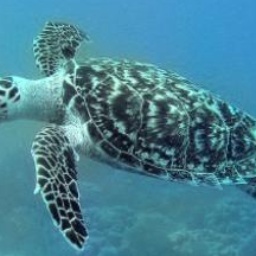}
        \end{minipage}
        \begin{minipage}[b]{.12\linewidth}
            \centering
            \includegraphics[width=\linewidth]{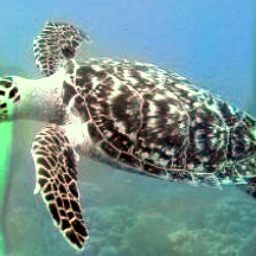}
        \end{minipage}
        \begin{minipage}[b]{.12\linewidth}
            \centering
            \includegraphics[width=\linewidth]{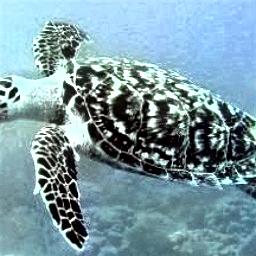}
        \end{minipage}
        \begin{minipage}[b]{.12\linewidth}
            \centering
            \includegraphics[width=\linewidth]{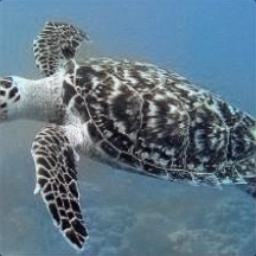}
        \end{minipage}
        \begin{minipage}[b]{.12\linewidth}
            \centering
            \includegraphics[width=\linewidth]{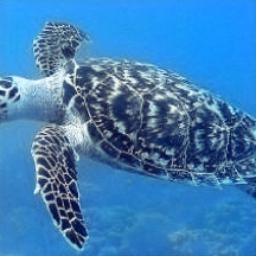}
        \end{minipage}
        \begin{minipage}[b]{.12\linewidth}
            \centering
            \includegraphics[width=\linewidth]{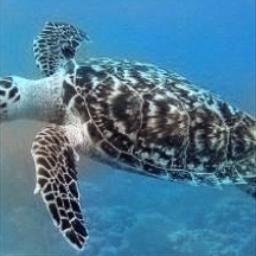}
        \end{minipage}
        \begin{minipage}[b]{.12\linewidth}
            \centering
            \includegraphics[width=\linewidth]{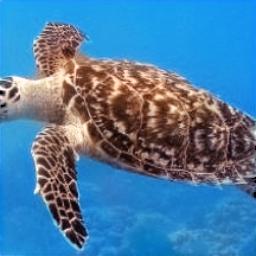}
        \end{minipage}
        \begin{minipage}[b]{.12\linewidth}
            \centering
            \includegraphics[width=\linewidth]{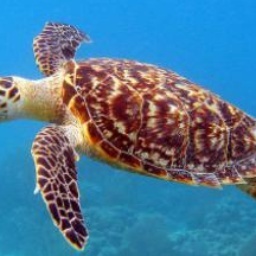}
        \end{minipage}
    \end{minipage}

    \begin{minipage}[b]{1\linewidth}
        \begin{minipage}[b]{.12\linewidth}
            \centering
            \includegraphics[width=\linewidth]{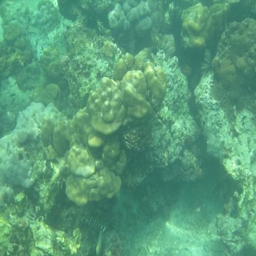}
        \end{minipage}
        \begin{minipage}[b]{.12\linewidth}
            \centering
            \includegraphics[width=\linewidth]{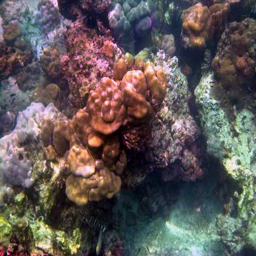}
        \end{minipage}
        \begin{minipage}[b]{.12\linewidth}
            \centering
            \includegraphics[width=\linewidth]{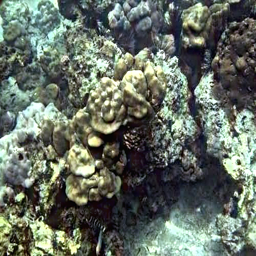}
        \end{minipage}
        \begin{minipage}[b]{.12\linewidth}
            \centering
            \includegraphics[width=\linewidth]{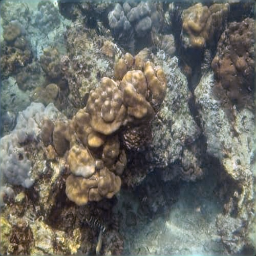}
        \end{minipage}
        \begin{minipage}[b]{.12\linewidth}
            \centering
            \includegraphics[width=\linewidth]{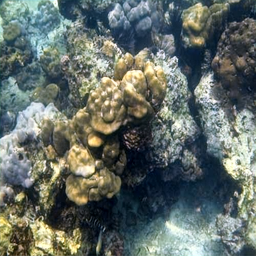}
        \end{minipage}
        \begin{minipage}[b]{.12\linewidth}
            \centering
            \includegraphics[width=\linewidth]{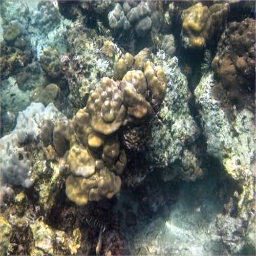}
        \end{minipage}
        \begin{minipage}[b]{.12\linewidth}
            \centering
            \includegraphics[width=\linewidth]{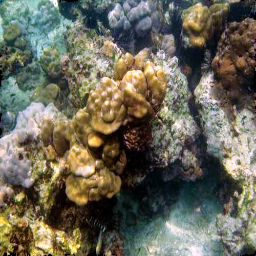}
        \end{minipage}
        \begin{minipage}[b]{.12\linewidth}
            \centering
            \includegraphics[width=\linewidth]{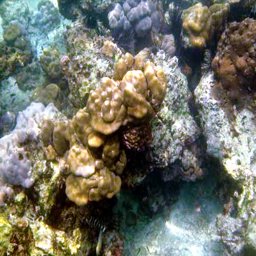}
        \end{minipage}
    \end{minipage}

    \begin{minipage}[b]{1\linewidth}
        \begin{minipage}[b]{.12\linewidth}
            \centering
            \centerline{\includegraphics[width=\linewidth]{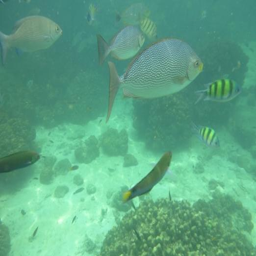}}
            \centerline{(a) Input}\medskip
        \end{minipage}
        \begin{minipage}[b]{.12\linewidth}
            \centering
            \centerline{\includegraphics[width=\linewidth]{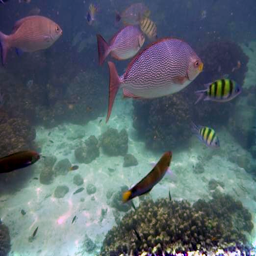}}
            \centerline{(b) ADPCC}\medskip
        \end{minipage}
        \begin{minipage}[b]{0.12\linewidth}
            \centering
            \centerline{\includegraphics[width=\linewidth]{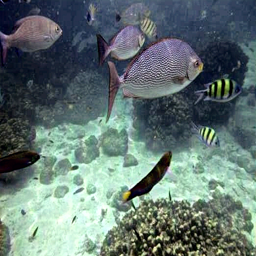}}
            \centerline{(c) WWPF}\medskip
        \end{minipage}
        \begin{minipage}[b]{.12\linewidth}
            \centering
            \centerline{\includegraphics[width=\linewidth]{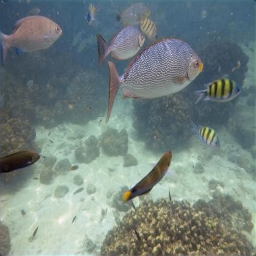}}
            \centerline{(d) GUPDM}\medskip
        \end{minipage}
        \begin{minipage}[b]{.12\linewidth}
            \centering
            \centerline{\includegraphics[width=\linewidth]{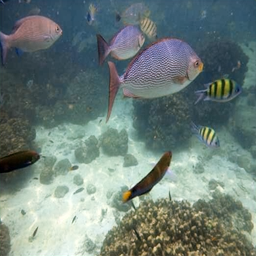}}
            \centerline{(e) Semi-UIR}\medskip
        \end{minipage}
        \begin{minipage}[b]{.12\linewidth}
            \centering
            \centerline{\includegraphics[width=\linewidth]{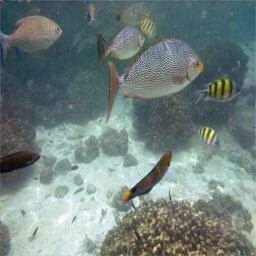}}
            \centerline{(f) URSCT}\medskip
        \end{minipage}
        \begin{minipage}[b]{.12\linewidth}
            \centering
            \centerline{\includegraphics[width=\linewidth]{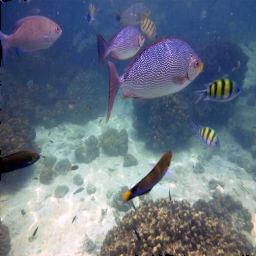}}
            \centerline{(g) Ours}\medskip
        \end{minipage}
        \begin{minipage}[b]{0.12\linewidth}
            \centering
            \centerline{\includegraphics[width=\linewidth]{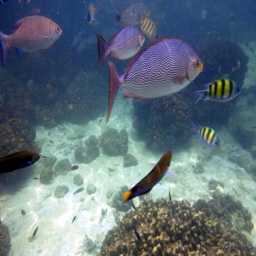}}
            \centerline{(h) Reference}\medskip
        \end{minipage}
    \end{minipage}
    
    \caption{Visual comparison of the proposed method with two leading traditional approaches and the three top-performing deep learning methods.} 
    \label{fig:comp}
\end{figure*}

%% file: tabletex/ablation.tex
\begin{table}[!b]
\caption{Ablation results averaged on UIEB, EUVP and LSUI.}
\label{tab:ab}
\adjustbox{width=\columnwidth}{
\begin{tabular}{ccccc|cccc}
\toprule
\multirow{2}{*}{UNet} & \multirow{2}{*}{CSC} & \multirow{2}{*}{LKA} & \multicolumn{2}{c|}{HDA} & \multirow{2}{*}{PSNR$\uparrow$} & \multirow{2}{*}{SSIM$\uparrow$} & \multirow{2}{*}{LPIPS$\downarrow$} & \multirow{2}{*}{UCIQE$\uparrow$} \\
                      &                      &                      & SDCA        & FDPA       &                       &                       &                        &                        \\ \midrule \checkmark 
                      & \ding{55}                    & \ding{55}                     & \ding{55}            & \ding{55}           & 23.13                 & 0.871                 & 0.173                  & 0.431                  \\
                    \checkmark  & \checkmark                     &  \ding{55}                    & \ding{55}            & \ding{55}           & 24.42                 & 0.882                 & 0.151                  & 0.439                  \\
                     \checkmark & \ding{55}                     & \checkmark                     & \ding{55}            & \ding{55}           & 24.71                 & 0.889                 & 0.137                  & 0.449                  \\
                     \checkmark & \checkmark                     &  \checkmark                    & \ding{55}            & \ding{55}           & 24.86                 & 0.888                 & 0.137                  & 0.445                  \\
                     \checkmark & \checkmark                     &  \checkmark                    & \checkmark            & \ding{55}           & 25.24                 & 0.891                 & 0.133                  & 0.450                  \\
                     \checkmark & \checkmark                     & \checkmark                     &  \checkmark           & \checkmark           & \textbf{25.50}                 & \textbf{0.893}                 & \textbf{0.128}                  & \textbf{0.453}                  \\ \bottomrule
\end{tabular}
}
\end{table}

%% file: IEEE-conference-template-062824.bbl
\begin{thebibliography}{10}
\providecommand{\url}[1]{#1}
\csname url@samestyle\endcsname
\providecommand{\newblock}{\relax}
\providecommand{\bibinfo}[2]{#2}
\providecommand{\BIBentrySTDinterwordspacing}{\spaceskip=0pt\relax}
\providecommand{\BIBentryALTinterwordstretchfactor}{4}
\providecommand{\BIBentryALTinterwordspacing}{\spaceskip=\fontdimen2\font plus
\BIBentryALTinterwordstretchfactor\fontdimen3\font minus \fontdimen4\font\relax}
\providecommand{\BIBforeignlanguage}[2]{{%
\expandafter\ifx\csname l@#1\endcsname\relax
\typeout{** WARNING: IEEEtran.bst: No hyphenation pattern has been}%
\typeout{** loaded for the language `#1'. Using the pattern for}%
\typeout{** the default language instead.}%
\else
\language=\csname l@#1\endcsname
\fi
#2}}
\providecommand{\BIBdecl}{\relax}
\BIBdecl

\bibitem{xie2021variational}
J.~Xie, G.~Hou, G.~Wang, and Z.~Pan, ``A variational framework for underwater image dehazing and deblurring,'' \emph{TCSVT}, vol.~32, no.~6, pp. 3514--3526, 2021.

\bibitem{zhang2022underwater}
W.~Zhang, P.~Zhuang, H.-H. Sun, G.~Li, S.~Kwong, and C.~Li, ``Underwater image enhancement via minimal color loss and locally adaptive contrast enhancement,'' \emph{TIP}, vol.~31, pp. 3997--4010, 2022.

\bibitem{liu2022rank}
J.~Liu, R.~W. Liu, J.~Sun, and T.~Zeng, ``Rank-one prior: Real-time scene recovery,'' \emph{TPAMI}, 2022.

\bibitem{zhou2023underwater}
J.~Zhou, Q.~Liu, Q.~Jiang, W.~Ren, K.-M. Lam, and W.~Zhang, ``Underwater camera: Improving visual perception via adaptive dark pixel prior and color correction,'' \emph{IJCV}, pp. 1--19, 2023.

\bibitem{zhang2023underwater}
W.~Zhang, L.~Zhou, P.~Zhuang, G.~Li, X.~Pan, W.~Zhao, and C.~Li, ``Underwater image enhancement via weighted wavelet visual perception fusion,'' \emph{TCSVT}, 2023.

\bibitem{zhang1}
X.~Zhang, F.~Chen, C.~Wang, M.~Tao, and G.-P. Jiang, ``Sienet: Siamese expansion network for image extrapolation,'' \emph{SPL}, vol.~27, pp. 1590--1594, 2020.

\bibitem{zhang6}
X.~Zhang, Z.~Xu, H.~Tang, C.~Gu, S.~Zhu, and X.~Guan, ``Shadclips: When parameter-efficient fine-tuning with multimodal meets shadow removal,'' 2024.

\bibitem{li2021underwater}
C.~Li, S.~Anwar, J.~Hou, R.~Cong, C.~Guo, and W.~Ren, ``Underwater image enhancement via medium transmission-guided multi-color space embedding,'' \emph{TIP}, vol.~30, pp. 4985--5000, 2021.

\bibitem{li2022beyond}
K.~Li, L.~Wu, Q.~Qi, W.~Liu, X.~Gao, L.~Zhou, and D.~Song, ``Beyond single reference for training: underwater image enhancement via comparative learning,'' \emph{TCSVT}, 2022.

\bibitem{zhao2024toward}
C.~Zhao, W.~Cai, C.~Dong, and Z.~Zeng, ``Toward sufficient spatial-frequency interaction for gradient-aware underwater image enhancement,'' in \emph{ICASSP}, 2024, pp. 3220--3224.

\bibitem{liu2022twin}
R.~Liu, Z.~Jiang, S.~Yang, and X.~Fan, ``Twin adversarial contrastive learning for underwater image enhancement and beyond,'' \emph{TIP}, vol.~31, pp. 4922--4936, 2022.

\bibitem{ma2022wavelet}
Z.~Ma and C.~Oh, ``A wavelet-based dual-stream network for underwater image enhancement,'' in \emph{ICASSP}, 2022, pp. 2769--2773.

\bibitem{cong2023pugan}
R.~Cong, W.~Yang, W.~Zhang, C.~Li, C.-L. Guo, Q.~Huang, and S.~Kwong, ``Pugan: Physical model-guided underwater image enhancement using gan with dual-discriminators,'' \emph{TIP}, 2023.

\bibitem{wang2023domain}
Z.~Wang, L.~Shen, M.~Xu, M.~Yu, K.~Wang, and Y.~Lin, ``Domain adaptation for underwater image enhancement,'' \emph{TIP}, vol.~32, pp. 1442--1457, 2023.

\bibitem{peng2023u}
L.~Peng, C.~Zhu, and L.~Bian, ``U-shape transformer for underwater image enhancement,'' \emph{TIP}, 2023.

\bibitem{ren2022reinforced}
T.~Ren, H.~Xu, G.~Jiang, M.~Yu, X.~Zhang, B.~Wang, and T.~Luo, ``Reinforced swin-convs transformer for simultaneous underwater sensing scene image enhancement and super-resolution,'' \emph{IEEE TGRS}, vol.~60, pp. 1--16, 2022.

\bibitem{huang2023contrastive}
S.~Huang, K.~Wang, H.~Liu, J.~Chen, and Y.~Li, ``Contrastive semi-supervised learning for underwater image restoration via reliable bank,'' in \emph{CVPR}, 2023, pp. 18\,145--18\,155.

\bibitem{fu2022unsupervised}
Z.~Fu, H.~Lin, Y.~Yang, S.~Chai, L.~Sun, Y.~Huang, and X.~Ding, ``Unsupervised underwater image restoration: From a homology perspective,'' in \emph{AAAI}, vol.~36, 2022, pp. 643--651.

\bibitem{mu2023generalized}
P.~Mu, H.~Xu, Z.~Liu, Z.~Wang, S.~Chan, and C.~Bai, ``A generalized physical-knowledge-guided dynamic model for underwater image enhancement,'' in \emph{ACM MM}, 2023, pp. 7111--7120.

\bibitem{zhu2024test}
L.~Zhu, W.~Liu, X.~Chen, Z.~Li, X.~Chen, Z.~Wang, and C.-M. Pun, ``Test-time intensity consistency adaptation for shadow detection,'' \emph{arXiv}, 2024.

\bibitem{xu2024muraldiff}
Z.~Xu, X.~Zhang, W.~Chen, J.~Liu, T.~Xu, and Z.~Wang, ``Muraldiff: Diffusion for ancient murals restoration on large-scale pre-training,'' \emph{TETCI}, 2024.

\bibitem{zhang8}
X.~Zhang, C.~Shen, X.~Yuan, S.~Yan, L.~Xie, W.~Wang, C.~Gu, H.~Tang, and J.~Ye, ``From redundancy to relevance: Enhancing explainability in multimodal large language models,'' \emph{arXiv}, 2024.

\bibitem{li2024cross}
X.~Li, G.~Huang, L.~Cheng, G.~Zhong, W.~Liu, X.~Chen, and M.~Cai, ``Cross-domain visual prompting with spatial proximity knowledge distillation for histological image classification,'' \emph{Journal of Biomedical Informatics}, vol. 158, p. 104728, 2024.

\bibitem{he2024residual}
Y.~He, W.~Song, L.~Li, T.~Zhan, and W.~Jiao, ``Residual feature-reutilization inception network,'' \emph{PR}, vol. 152, p. 110439, 2024.

\bibitem{li2022nndf}
L.~Li, Y.~He, and L.~Li, ``Nndf: A new neural detection network for aspect-category sentiment analysis,'' in \emph{KSEM}, 2022, pp. 339--355.

\bibitem{zhang2}
X.~Zhang, Y.~Zhao, C.~Gu, C.~Lu, and S.~Zhu, ``Spa-former: An effective and lightweight transformer for image shadow removal,'' in \emph{IJCNN}, 2023, pp. 1--8.

\bibitem{liu2024forgeryttt}
W.~Liu, X.~Shen, C.-M. Pun, and X.~Cun, ``Forgeryttt: Zero-shot image manipulation localization with test-time training,'' \emph{arXiv}, 2024.

\bibitem{li2024high-fidelity}
M.~Li, H.~Sun, Y.~Lei, X.~Zhang, Y.~Dong, Y.~Zhou, Z.~Li, and X.~Chen, ``High-fidelity document stain removal via a large-scale real-world dataset and a memory-augmented transformer,'' in \emph{WACV}, 2024.

\bibitem{zhang2024seeing}
X.~Zhang, Y.~Quan, C.~Gu, C.~Shen, X.~Yuan, S.~Yan, H.~Cheng, K.~Wu, and J.~Ye, ``Seeing clearly by layer two: Enhancing attention heads to alleviate hallucination in lvlms,'' \emph{arXiv}, 2024.

\bibitem{liu2024dh}
W.~Liu, X.~Cun, and C.-M. Pun, ``Dh-gan: Image manipulation localization via a dual homology-aware generative adversarial network,'' \emph{PR}, p. 110658, 2024.

\bibitem{li2023high-resolution}
Z.~Li, X.~Chen, C.-M. Pun, and X.~Cun, ``High-resolution document shadow removal via a large-scale real-world dataset and a frequency-aware shadow erasing net,'' in \emph{ICCV}, 2023, pp. 12\,449--12\,458.

\bibitem{guo2024dual-hybrid}
X.~Guo, X.~Chen, S.~Luo, S.~Wang, and C.-M. Pun, ``Dual-hybrid attention network for specular highlight removal,'' in \emph{ACM MM}, 2024, pp. 10\,173--10\,181.

\bibitem{liu2023explicit}
W.~Liu, X.~Shen, C.-M. Pun, and X.~Cun, ``Explicit visual prompting for low-level structure segmentations,'' in \emph{CVPR}, 2023, pp. 19\,434--19\,445.

\bibitem{zhang10}
X.~Zhang, H.~Tang, C.~Gu, and S.~Zhu, ``Enlighten-anything: When segment anything model meets low-light image enhancement,'' \emph{arXiv preprint arXiv:2306.10286}, 2023.

\bibitem{zhang11}
X.~Mei, X.~Ye, X.~Zhang, Y.~Liu, J.~Wang, J.~Hou, and X.~Wang, ``Uir-net: a simple and effective baseline for underwater image restoration and enhancement,'' \emph{Remote Sensing}, vol.~15, no.~1, p.~39, 2022.

\bibitem{yuan2024instance}
X.~Yuan, C.~Shen, S.~Yan, X.~Zhang, L.~Xie, W.~Wang, R.~Guan, Y.~Wang, and J.~Ye, ``Instance-adaptive zero-shot chain-of-thought prompting,'' \emph{arXiv}, 2024.

\bibitem{zhang9}
J.~Wei and X.~Zhang, ``Dopra: Decoding over-accumulation penalization and re-allocation in specific weighting layer,'' \emph{arXiv}, 2024.

\bibitem{liu2023coordfill}
W.~Liu, X.~Cun, C.-M. Pun, M.~Xia, Y.~Zhang, and J.~Wang, ``Coordfill: Efficient high-resolution image inpainting via parameterized coordinate querying,'' in \emph{AAAI}, vol.~37, no.~2, 2023, pp. 1746--1754.

\bibitem{liu2024depth}
W.~Liu, X.~Shen, H.~Li, X.~Bi, B.~Liu, C.-M. Pun, and X.~Cun, ``Depth-aware test-time training for zero-shot video object segmentation,'' in \emph{CVPR}, 2024, pp. 19\,218--19\,227.

\bibitem{he2024generalized}
Y.~He, L.~Li, T.~Zhan, W.~Jiao, and C.-M. Pun, ``Generalized uncertainty-based evidential fusion with hybrid multi-head attention for weak-supervised temporal action localization,'' in \emph{ICASSP}, 2024, pp. 3855--3859.

\bibitem{guo2023visual}
M.-H. Guo, C.-Z. Lu, Z.-N. Liu, M.-M. Cheng, and S.-M. Hu, ``Visual attention network,'' \emph{cvm}, vol.~9, no.~4, pp. 733--752, 2023.

\bibitem{ding2022scaling}
X.~Ding, X.~Zhang, J.~Han, and G.~Ding, ``Scaling up your kernels to 31x31: Revisiting large kernel design in cnns,'' in \emph{CVPR}, 2022, pp. 11\,963--11\,975.

\bibitem{cui2024omni}
Y.~Cui, W.~Ren, and A.~Knoll, ``Omni-kernel network for image restoration,'' in \emph{AAAI}, vol.~38, no.~2, 2024, pp. 1426--1434.

\bibitem{lau2024large}
K.~W. Lau, L.-M. Po, and Y.~A.~U. Rehman, ``Large separable kernel attention: Rethinking the large kernel attention design in cnn,'' \emph{ESWA}, vol. 236, p. 121352, 2024.

\bibitem{li2019underwater}
C.~Li, C.~Guo, W.~Ren, R.~Cong, J.~Hou, S.~Kwong, and D.~Tao, ``An underwater image enhancement benchmark dataset and beyond,'' \emph{TIP}, vol.~29, pp. 4376--4389, 2019.

\bibitem{islam2020fast}
M.~J. Islam, Y.~Xia, and J.~Sattar, ``Fast underwater image enhancement for improved visual perception,'' \emph{IEEE RAL}, vol.~5, no.~2, pp. 3227--3234, 2020.

\bibitem{liu2020real}
R.~Liu, X.~Fan, M.~Zhu, M.~Hou, and Z.~Luo, ``Real-world underwater enhancement: Challenges, benchmarks, and solutions under natural light,'' \emph{TCSVT}, vol.~30, no.~12, pp. 4861--4875, 2020.

\end{thebibliography}
